\documentclass{article}

\PassOptionsToPackage{numbers}{natbib}

\usepackage{amsmath}
\usepackage{graphicx}
\usepackage{subcaption}
\usepackage{tabularx}
\usepackage{float} % For improved figure placement
\usepackage{makecell}
\usepackage{array}
\usepackage{multirow} 
\usepackage{bbm}

\usepackage[final]{neurips_2024}

\usepackage[utf8]{inputenc} % allow utf-8 input
\usepackage[T1]{fontenc}    % use 8-bit T1 fonts
\usepackage{hyperref}       % hyperlinks
\usepackage{url}            % simple URL typesetting
\usepackage{booktabs}       % professional-quality tables
\usepackage{amsfonts}       % blackboard math symbols
\usepackage{nicefrac}       % compact symbols for 1/2, etc.
\usepackage{microtype}      % microtypography
\usepackage{xcolor}         % colors
\definecolor{darkgreen}{rgb}{0.0, 0.5, 0.0}

\title{Evaluating Fine-Tuning Efficiency of Human-Inspired Learning Strategies in Medical Question Answering}
\author{
    Yushi Yang\thanks{Correspondence: Yushi Yang, 
    <yushi.yang@oii.ox.ac.uk>}, \, Andrew M. Bean, \, Robert McCraith, \,  Adam Mahdi \vspace{1ex} \\
    University of Oxford
}

\begin{document}

\maketitle

% \begin{center}
% \vspace{-0.8cm}
% {\bf University of Oxford}
% \end{center}

\begin{abstract}
% background
Fine-tuning Large Language Models (LLMs) incurs considerable training costs, driving the need for data-efficient training with optimised data ordering. Human-inspired strategies offer a solution by organising data based on human learning practices.
% method
This study evaluates the fine-tuning efficiency of five human-inspired strategies across four language models, three datasets, and both human- and LLM-labelled data in the context of medical question answering.
%this study evaluates the efficiency of human-inspired learning strategies for fine-tuning across different language models, datasets, and data labels. Our study rigorously assesses 5 learning strategies across 4 models, 3 datasets, and both human- and LLM-labelled data.
% We rigorously assess five strategies, including non-curriculum-based approaches, across three datasets and four models, with both human- and LLM-labelled data.
% we evaluate five human-inspired learning strategies for fine-tuning four LLMs, comparing both human-defined and LLM-defined data labels on question difficulty and categories.
% results and discussion
These strategies achieve the best accuracy gain of 1.81\% and an average gain of 1.02\% across datasets, with interleaved strategies delivering the best average results. However, the best strategy varies across model-dataset combinations, limiting the generalisability of the effects of any single strategy. Additionally, LLM-defined question difficulty outperforms human-defined labels in curriculum-based learning, showing the potential of model-generated data as a cost-effective alternative for optimising fine-tuning. \footnote{The code is available at: \href{https://github.com/Oxford-AI-for-Society/human-learning-strategies/}{https://github.com/Oxford-AI-for-Society/human-learning-strategies/}.}
% (averaged across models)
% (averaged across datasets)
%  the best accuracy gains of 1.81\% across datasets and 1.44\% across models,
\end{abstract}

%%%%%%%%%%%%%%%%%%%%%%%%%%%%%%%%%%%%%%%
\section{Introduction}
%%%%% wider context to niche
Training large language models (LLMs) is costly, both in terms of data collection efforts \cite{llm_data_repetition, llm_data_repetition_2} and computational resources \cite{llm_data_expensive, llm_data_expensive_2}. To mitigate these costs, recent research has focused on data-efficient training methods that optimise data ordering \cite{llm_data_efficient_pretrain}. One promising approach is human-inspired learning strategies, which organise data based on human learning principles \cite{zhang2023learningpredictconceptordering, curri_nlu}. Among these strategies, curriculum learning, where samples are arranged from easiest to hardest, is particularly popular \cite{curri_easy_good, curri_nlu}. This method has proven effective in language tasks such as knowledge acquisition \cite{llm_curriculum_learning}, natural language reasoning \cite{curri_reasoning}, and information retrieval \cite{curri_ir}.

% To mitigate these costs, recent studies have investigated data-efficient training methods inspired by human learning for efficient data ordering and selection \cite{curri_easy_good, llm_data_efficient_pretrain}. The most well-known of these methods is curriculum learning, where training samples are ordered from easiest to hardest to mirror human learning patterns \cite{curri_easy_good, curri_nlu}. 

%%%%%% Research/literature gap
While curriculum learning has shown promise, the broader potential of human-inspired data ordering for fine-tuning remains unclear. Specifically, previous studies have focused on (i) a narrower set of models \cite{curri_reasoning, curri_nlu}, with fewer comparisons across multiple models; (ii) limited evaluation in domain-specific applications, especially in high-stakes areas like medical question answering; and (iii) primarily using one set of data labels, with less emphasis on comparing human- versus machine-defined labels to reflect different perspectives on data ordering.
% (i) limited exploration of learning strategies beyond curriculum learning; 
To address these gaps, we conduct a comprehensive evaluation of human-inspired learning strategies for fine-tuning in medical question answering, testing models of various sizes using both human- and LLM-defined question difficulty labels. We focus on the medical question-answering domain because of the scarcity of high-quality medical questions needed to train effective medical LLMs, highlighting the importance of efficient fine-tuning strategies \cite{lehman2023needclinicallanguagemodels, manathunga2023aligninglargelanguagemodels}. Our findings offer empirical insights into the efficiency and generalisability of these strategies for fine-tuning LLMs. Specifically, we find that:
%  such as a comparison between human- and machine-labelled sample difficulty for the effectiveness of curriculum learning.
% several language models fine-tuned on medical data have recently been released to enhance accurate information retrieval in the medical context \cite{med_gemini, meditron, medpalm2}.

%  On the other hand, previous studies on curriculum learning have primarily focused on a single model with a specific curriculum strategy \cite{curri_reasoning, curri_nlu}, offering valuable insights but leaving questions about the generalisability of the results. 

%%%%% Contributions
\begin{itemize}
    % \item \textbf{Broad-based evaluation of human-inspired learning strategies for fine-tuning}: Unlike previous work that focused on individual language model and curriculum learning, we extended the analysis to four LLMs of three different sizes and architectures and non-curriculum-based learning strategies. 
    \item \textit{Human-inspired learning yields moderate improvements over random shuffling.} These strategies result in the best accuracy gain of 1.81\% and an average gain of 1.02\% across datasets, with interleaved strategies providing the best average results.
    %consistent with results in prior studies \cite{curri_nlu, curri_reasoning}.
    \item \textit{Human-inspired learning lacks generalisation across model-data combinations.} 
    The best strategy varies across model-dataset combinations, suggesting caution when generalising the effects of any one strategy to other models based on single-model results.
    % The effectiveness of human-inspired learning strategies varies significantly across model-dataset combinations, indicating that a strategy effective for one model may not generalise to others.
    \item \textit{LLM-defined difficulty outperforms human labels in curriculum-based learning.} We automatically labelled question difficulty using an ensemble of LLM responses. The results show that switching to LLM-defined difficulty modestly improves the performance of curriculum-based strategies, offering a cost-effective alternative to human annotations for optimising fine-tuning.
    % and categories  and text clustering
   % This method leverages pre-trained LLM knowledge for data labelling, offering a cost-effective alternative to human annotations. We find that using LLM-defined question difficulty improves curriculum learning performance compared to human-defined labels.
    % Comparison of machine-generated and human-generated data labels
\end{itemize}

% key graph - learning orders
\begin{figure}[t!]
    \centering
    \includegraphics[width=1.0\textwidth]{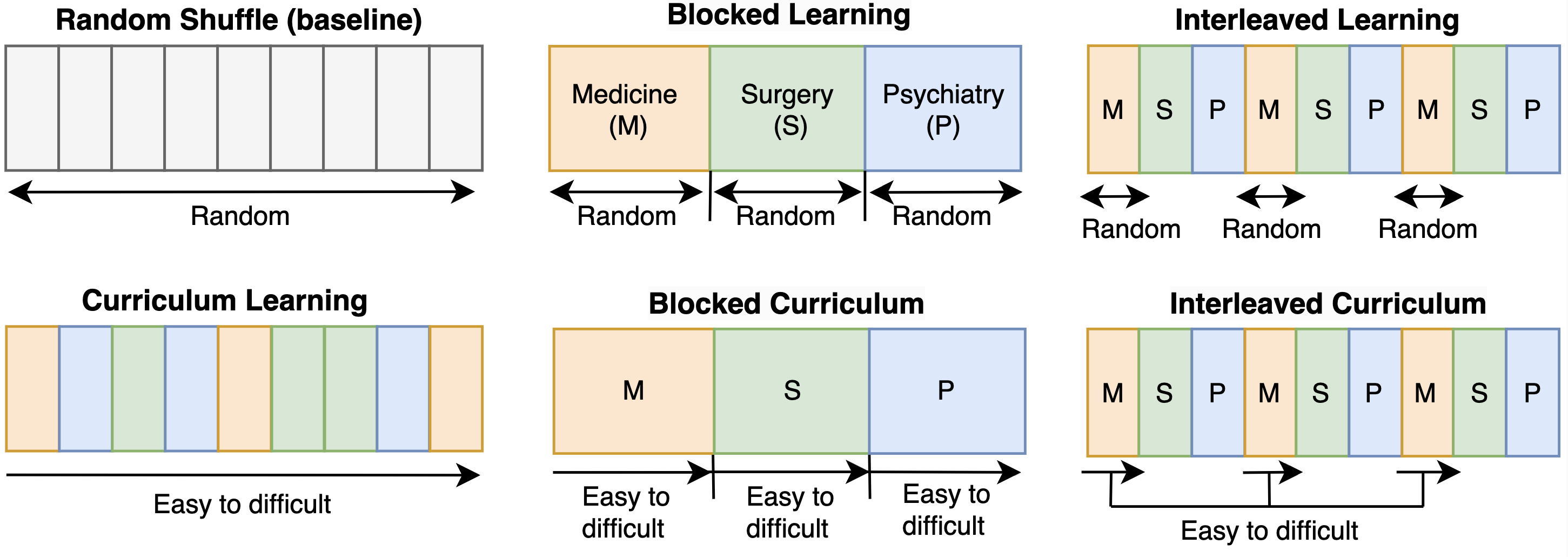}
    \caption{\textbf{Human-inspired learning strategies.} Along the \textit{Random Shuffle} baseline, five human-inspired learning strategies order data based on question difficulty (shown with arrows) and question category (represented by block colors). The first row shows non-curriculum-based strategies, while the second row shows curriculum-based strategies.}
    %:(i) \textit{Blocked Learning:} Questions are grouped by category and randomised within each group. (ii) \textit{Interleaved Learning:} Questions are grouped by category, then each category is randomly divided into three equal parts, and questions from each part are arranged in an interleaved manner. (iii) \textit{Curriculum Learning:} Questions are sorted by difficulty in ascending order. (iv) \textit{Blocked Curriculum:} Questions are grouped by category and then arranged in ascending difficulty within each category. (v) \textit{Interleaved Curriculum:} Following Blocked Curriculum, questions in each category are further divided into three equal parts, and then interleaved.}
    \label{fig:learning_orders}
\end{figure}

%%%%%%%%%%%%%%%%%%%%%%%%%%%%%%%%%%%%%%%
\section{Experimental design}
We evaluated the efficiency of human-inspired data ordering strategies for fine-tuning LLMs in medical question answering. Our study compared 5 human-inspired learning strategies against a Random Shuffle baseline  (Section \ref{sec:learning_strategies}) across 4 LLMs (Section \ref{sec:models}) and 3 medical datasets (Section \ref{sec:datasets}). We also applied these strategies with both human- and model-generated labels on question difficulty, creating 2 data-labelling scenarios (Section \ref{sec:data_labelling}). 
This results in 144 fine-tuned models (144 = 6 strategies × 4 models × 3 datasets × 2 data labels).
% This results in 216 fine-tuned models (216 = 6 strategies × 4 models × 3 datasets × 3 data labels).

% \footnote{Our code for the fine-tuning experiments can be found at \href{https://github.com/Oxford-AI-for-Society/human-learning-strategies}{https://github.com/Oxford-AI-for-Society/human-learning-strategies}.}

\subsection{Human-inspired learning strategies}
% \paragraph{Human-inspired learning strategies}
\label{sec:learning_strategies}
Figure \ref{fig:learning_orders} presents five human-inspired learning strategies. These strategies are defined based on two data labels: (i) a continuous measure of \textit{question difficulty} and (ii) a discrete measure of \textit{question category}. Among them, Blocked Learning and Interleaved Learning rely solely on question category and do not follow a curriculum. The remaining strategies use the difficulty measure to create a curriculum, ranking questions from easiest to hardest.

% Inspired by Lee et al. \cite{llm_curriculum_learning}, who integrated blocking and interleaving practices into curriculum design, we modified these strategies by strictly sorting questions based on continuous difficulty values. This approach avoids arbitrary distinctions between questions of similar difficulty, unlike the original easy, medium, and hard categorisations.

%\paragraph{Motivations from human learning}
% can potentially improve the effectiveness of LLMs by structuring their learning process to promote better memory retention, generalisation, and prevent catastrophic forgetting \cite{catastrophic_forgetting}. 
The five learning strategies are based on common educational practices for human learning. \textit{Blocked Learning} groups questions by category, like blocked practice in education \cite{blocked_1, blocked_2}, where focusing on one topic deepens understanding. \textit{Interleaved Learning} alternates questions from different categories, revisiting them periodically to reduce cognitive decay and improve retention \cite{interleaved_1, interleaved_2}. \textit{Curriculum Learning} orders questions from easiest to hardest by question difficulty, akin to traditional curricula where basic knowledge precedes complex tasks \cite{curriculum_learning}. \textit{Blocked Curriculum} combines blocked and curriculum learning, arranging questions by difficulty within each category to progressively build knowledge \cite{llm_curriculum_learning}. \textit{Interleaved Curriculum} cycles through categories with increasing question difficulty, reinforcing learning by revisiting topics with harder material \cite{spiral_curri}.

% Finally, Adaptive Learning begins with a mix of domains and difficulties, and as the model demonstrates proficiency in certain areas, increase the difficulty or switch focus to less well-mastered domains. This simulates personalised learning trajectories based on performance feedback and reinforce on questions that are more likely to get wrong.

\subsection{Language models} 
\label{sec:models}
% This selection measures the effects of learning strategies across three varying sizes and two different model architectures. 
We fine-tuned four open-source language models of varying sizes and architectures: TinyLlama 1.1B \cite{tinyllama}, Llama 2 7B \cite{llama2}, Llama 2 13B \cite{llama2}, and Mistral 7B \cite{mistral}. TinyLlama shares the architecture of Llama 2 with only 1.1B parameters \cite{tinyllama}. For the two Llama 2 models, we used the chat versions optimised for dialogue, focusing on fine-tuning medical knowledge rather than instruction styles. All models were accessed via Hugging Face and fine-tuned on two NVIDIA RTX 6000 Ada cards. To optimise memory, the models were quantised to 4-bits with double quantisation. 
% We excluded the 70B models due to the high computational cost of repeated fine-tuning.
% and tokenizer 

\subsection{Datasets}
\label{sec:datasets} 
We fine-tuned on a single medical question answering dataset with human-defined labels on question difficulty and categories, then evaluated on three datasets to test the models' generalisation. For fine-tuning, we used the Lekarski Egzamin Końcowy (LEK) dataset \cite{our_lek}, consisting of multiple-choice questions from Polish medical licensing exams\footnote{The LEK dataset is publicly available at \url{https://cem.edu.pl/}}.  Unlike other medical exam datasets, LEK includes test-takers' responses for each question, allowing us to assess question difficulty based on actual human performance. We used the English version of questions from the last five exams (spring 2021 to spring 2023), resulting in a final dataset of 874 unique questions across ten medical categories. For evaluation, we used (i) the LEK test set with cross-validation, (ii) the official MedMCQA validation set \cite{medmcqa}\footnote{We used the validation set for evaluation following \citet{pmcllama} and \citet{meditron}, as the test set does not publicly provide answer keys.}, and (iii) the official MedQA test set \cite{medqa}.
% following \citet{pmcllama} and \citet{meditron}
% \footnote{We used the validation set for evaluation, following \citet{pmcllama} and \citet{meditron}, as the test set does not publicly provide answer keys.}

\subsection{Supervised fine-tuning}
% \paragraph{Instruction prompt}
\label{sec:instruction_prompt}
We used zero-shot instruction prompting for each question with the following instruction: \textit{Answer the following multiple-choice question by giving the most appropriate response. The answer should be one of [A, B, C, D, E].}
The instruction was then followed by the multiple choice question. The response template began with \textit{`Answer:'}, and during fine-tuning, the model learned to predict the correct answer index as the next token. Our zero-shot prompt structure is designed to reflect typical exam instructions and serves as a baseline for performance. The same prompt format was used during inference, where the correct answer index was masked and predicted.
% While more optimised prompts can be used for improving performance \cite{microsoft_fancy_prompting, medpalm2}
% We also experiemented with fine-tuning both the choice index and the answer text and found that has negligible effects on performance.

%\paragraph{Fine-tuning method} 
We used the QLoRA \cite{qlora} method for parameter-efficient fine-tuning on the linear layers of the language models. To preserve the specified learning order in Figure \ref{fig:learning_orders}, we disabled automatic data shuffling when loading data and trained the entire sequence in a single epoch. Training over multiple epochs would disrupt the learning order specified; for example, repeating Blocked practice multiple times would effectively turn it into an Interleaved practice, and re-running a curriculum would cause jumps from difficult to easy questions. We also experimented with repeating samples three times within each category to increase learning data, which yielded similar performance to no repetition. The fine-tuning hyperparameters, selected via grid search, are provided in Appendix \ref{appendix:hyperparameters}. 
% To ensure fair comparisons across learning strategies, all learning strategies applied to a model used the same hyperparameters as the Random Shuffle baseline.
% while maintaining the curriculum order

\subsection{Data labelling scenarios}
\label{sec:data_labelling}
We evaluated the effectiveness of learning strategies in two data labelling scenarios: with (a) human-defined question difficulty and (b) LLM-defined question difficulty. Human-defined difficulty is calculated as the proportion of incorrect answers from human test takers for each question in the LEK dataset, with a higher error rate indicating more difficult questions. We tested LLM-defined difficulty to extend learning strategies to unlabelled data, where human annotations are costly, which is especially true in high-stakes fields such as medical question answering.
% , with pre-existing labels used for question categories.

% using human- and machine-defined labels: 
% \begin{flushleft}
% \hspace{1cm} (a) Human-defined question difficulty and categories; \\
% \hspace{1cm} (b) LLM-defined question difficulty with human-defined categories; \\
% \hspace{1cm} (c) LLM-defined question difficulty with categories identified through text clustering.
% \end{flushleft}

% \begin{itemize} 
% \item (a) Human-defined question difficulty and categories; 
% \item (b) LLM-defined question difficulty with human-defined categories; 
% \item (c) LLM-defined question difficulty with categories identified through text clustering. \end{itemize}

% Human-defined labels refer to the pre-existing labels in the LEK dataset. Automated labels were tested to extend learning strategies to unlabelled data, where obtaining human annotations is costly.
% Details of the automated labelling process are described below.

%\paragraph{LLM-defined question difficulty}
We prompted six LLMs to answer the questions in the LEK dataset, using the instruction prompt in Section  \ref{sec:instruction_prompt}. For each LLM, we calculated an \textit{expected accuracy} for each question as the probability the LLM assigns to the correct answer during next-token prediction:
\begin{equation}\label{eq:example}
  \mathbb{E}[\text{Accuracy}] = \sum_{c \in \{A, B, C, D, E\}} P(c) \cdot \mathbbm{1}(c = c^*),
\end{equation}
where \( P(c) \) is the probability assigned to answer index \( c \in \{A, B, C, D, E\}\), and \( \mathbbm{1}(c = c^*) \) is 1 if \( c \) is the correct answer \( c^* \), otherwise 0. The LLM-defined difficulty for each question is defined as \textit{(1 - expected accuracy)}, averaged across the LLMs. The models used to compute difficulty include GPT-4 Turbo \cite{gpt4}, GPT-3.5 \cite{gpt3.5}, PaLM 2 \cite{palm2}, Mixtral 8x7B \cite{mixtral},  Llama 2 70B \cite{llama2}, and Meditron 70B \cite{meditron}, covering a wide range of architectures and pre-training medical knowledge. These models differ from those used for fine-tuning to ensure an unbiased assessment of question difficulty. They showed reasonable per-question agreement with an average correlation of 0.54.
% Results for labelling difficulty in the MedQA dataset using the ensemble models are provided in Appendix \ref{appendix:medqa_ft_results}.
% leveraging knowledge aggregated from a diverse set of models.

% We assume that averaging responses from LLMs with popular architectures offers an objective estimate of the general medical knowledge that LLMs possess. 

\subsection{Evaluation metrics}
\label{sec:evaluation}
We used greedy decoding to generate the model's answer index and compared it to the correct answer to calculate the \textit{accuracy score}. For each model, accuracy scores for a learning strategy were averaged across all datasets, and for each dataset, scores were averaged across all models (Table \ref{tab:all_combined_accuracy}). For each model or dataset, the \textit{accuracy gain} is measured as the difference between the best-performing strategy and the Random Shuffle baseline. The \textit{mean accuracy gain} of a strategy is measured as its accuracy gains averaged across all model-dataset combinations (the `Mean' column in Table \ref{tab:all_combined_accuracy}). Accuracy scores for each strategy across each model-dataset combination were averaged over five runs, with question category orders in Blocked and Interleaved Learning kept consistent across runs.

% the \textit{accuracy gain} as the difference in accuracy between a learning strategy and the Random Shuffle baseline. For a model or dataset

\renewcommand{\arraystretch}{1.12} % Increase line spacing 
\setlength{\tabcolsep}{4pt} % Adjust the horizontal padding between columns
\captionsetup[table]{skip=9pt} % Increase the distance between end of table and the caption
\begin{table*}[t!]
\centering
\fontsize{9.5}{10.5}\selectfont % Set custom font size (11pt) and line height (12pt)
% The learning strategies marked in grey in Tables (b) and (c) indicate unchanged results from Table (a) due to unchanged data labels. 
% Abbreviations: \textit{Blocked Curri.} = Blocked Curriculum, \textit{Interleaved Curri.} = Interleaved Curriculum, \textit{AVG} = average.
\caption{\textbf{The accuracy scores (in \%) for learning strategies averaged across datasets or models.} This table shows that the best strategy varies across models and datasets. In the \textit{Models} columns, scores are averaged over three datasets; in the \textit{Datasets} columns, scores are averaged over four models. The \textit{Mean} column shows the accuracy scores for a learning strategy averaged across all model-dataset combinations. Note that here \textit{Tiny} stands for TinyLlama.} 
\label{tab:all_combined_accuracy}
\resizebox{1\textwidth}{!}{ % adjust table horizontal length
\begin{minipage}{\textwidth}
    \centering
    \begin{subtable}{\textwidth}
        \centering
        \begin{tabularx}{0.97\textwidth}{|l|p{1.1cm}p{1.1cm}p{1.1cm}p{1.1cm}|p{1.1cm}p{1.1cm}p{1.0cm}|p{1.0cm}|}
        % \begin{tabularx}{1\textwidth}{|l|XXXX|XXX|}
    \hline
   \multicolumn{1}{|c|}{\textbf{Strategy}} & \multicolumn{4}{c|}{\textbf{Models}} & \multicolumn{3}{c|}{\textbf{Datasets}} & \multicolumn{1}{c|}{\textbf{Mean}} \\ 
    & \textbf{Tiny 1.1B} & \textbf{Llama\,2 7B} & \textbf{Llama\,2 13B} & \textbf{Mistral 7B} & \textbf{LEK} & \textbf{Med MCQA} & \textbf{Med QA} & \\ \hline
Random Shuffle         & 20.40  & 38.71 & 42.57 & 47.97 & 43.55 & 36.28 & 32.40 & 37.41 \\
Curriculum             & 19.79 & \textbf{39.05} & \textbf{43.68} & 47.31 & \textbf{44.68} & 36.36 & 31.35 & 37.46 \\ 
Blocked                & 20.47 & 38.46 & 42.83 & 48.10 & 43.99 & 36.45 & 31.97 & 37.47 \\ 
Blocked Curri.         & \textbf{21.80} & 38.32 & 42.57 & 47.10 & 43.84 & 36.46 & 32.05 & 37.45 \\ 
Interleaved            & 21.74 & 38.87 & 42.79 & \textbf{48.88} & 44.18 & \textbf{37.04} & \textbf{32.99} & \textbf{38.07} \\ 
Interleaved Curri.     & 21.10 & 38.10 & 42.69 & 48.04 & 43.81 & 36.44 & 32.20 & 37.48 \\ \hline
        \end{tabularx}
        \caption{Data label: human-defined difficulty}
        \label{tab:accuracy_human_human}
    \end{subtable}
    \vspace{0.01cm}

    \begin{subtable}{\textwidth}
        \centering
        \begin{tabularx}{0.97\textwidth}{|l|p{1.1cm}p{1.1cm}p{1.1cm}p{1.1cm}|p{1.1cm}p{1.1cm}p{1.0cm}|p{1.0cm}|}
        % \begin{tabularx}{1\textwidth}{|l|XXXX|XXX|}
    \hline
    \multicolumn{1}{|c|}{\textbf{Strategy}} & \multicolumn{4}{c|}{\textbf{Models}} & \multicolumn{3}{c|}{\textbf{Datasets}} & \multicolumn{1}{c|}{\textbf{Mean}} \\ 
    & \textbf{Tiny 1.1B} & \textbf{Llama\,2 7B} & \textbf{Llama\,2 13B} & \textbf{Mistral 7B} & \textbf{LEK} & \textbf{Med MCQA} & \textbf{Med QA} & \\ \hline
Random Shuffle         & 20.40 & 38.71 & 42.57 & 47.97 & 43.55 & 36.28 & 32.40 & 37.41 \\
Curriculum             & 20.88 & \textbf{39.21} & 42.82 & 48.39 & \textbf{44.36} & 36.86 & 32.26 & 37.83 \\ 
Blocked                & 20.47 & 38.46 & 42.83 & 48.10 & 43.99 & 36.45 & 31.97 & 37.47 \\ 
Blocked Curri.         & \textbf{21.84} & 37.89 & 42.67 & 48.71 & 43.64 & 37.20 & 32.51 & 37.78 \\ 
Interleaved            & 21.74 & 38.87 & 42.79 & 48.88 & 44.18 & 37.04 & \textbf{32.99} & 38.07 \\ 
Interleaved Curri.     & 21.67 & 38.98 & \textbf{43.02} & \textbf{49.32} & 44.22 & \textbf{38.09} & 32.43 & \textbf{38.25} \\ \hline
        \end{tabularx}
        \caption{Data label: LLM-defined difficulty} 
        \label{tab:accuracy_llm_human}
    \end{subtable}
\end{minipage}
}
\end{table*}

\captionsetup[figure]{skip=5pt} % Remove spacing between figure and caption
\captionsetup[table]{skip=5pt} % Remove spacing between table and caption
\begin{figure}[t!] % Force figure placement
    \centering
    \begin{subfigure}[t!]{0.49\textwidth} % Align minipage to the top
        \centering
        \vspace{-20pt} % Further reduce space above the figure
        \includegraphics[width=\textwidth, height=0.7\textwidth]{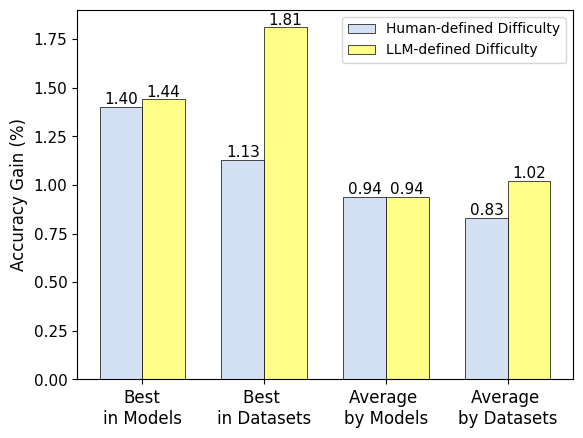} % Adjust figure height
        \caption{Effects of human-inspired learning strategies.} 
        \label{fig:best_accuracy_improvements}
    \end{subfigure}
    \hfill
    \begin{subfigure}[t!]{0.49\textwidth} % Align minipage to the top
        \centering
        \vspace{-20pt} % Further reduce space above the figure
        \includegraphics[width=\textwidth, height=0.7\textwidth]{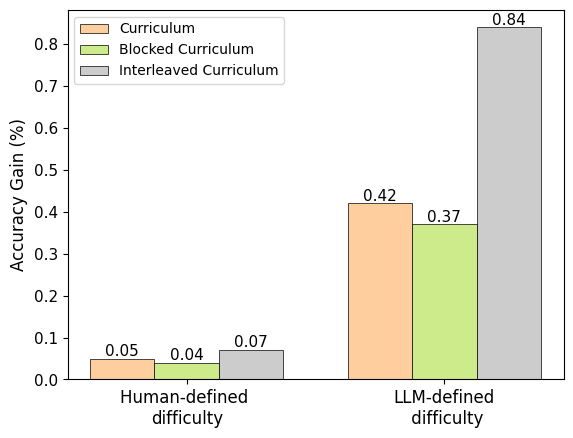} % Adjust figure height
        \caption{Accuracy gains for curriculum-based learning.}
        \label{fig:llm_diff_improvements}
    \end{subfigure}
    \caption{\textbf{Comparison of accuracy gains from learning strategies with human- and LLM-defined difficulty.} (a) shows the highest and average accuracy gains (in \%) of the best strategy across models and datasets, compared to the Random Shuffle baseline. (b) shows that using LLM-defined difficulty improves accuracy scores for all curriculum-based strategies, with each bar showing the mean accuracy gains (in \%) across all model-data combinations.}
    \label{fig:all_results}
\end{figure}

%%%%%%%%%%%%%%%%%%%%%%%%%%%%%%%%%%%%%%
\section{Results}
\paragraph{Impact of human-inspired learning strategies}
% Impact of fine-tuning using human-inspired learning strategies
\label{sec:results_1}
% RESULTS PARAGRAPH: short and to the point, no discussion
Overall, adopting a human-inspired learning strategy yields the best accuracy gain of 1.81\% and an average gain of 1.02\% across datasets (1.44\% and 0.94\% across models) over Random Shuffle (Figure \ref{fig:best_accuracy_improvements}), both significant at the 5\% level in a paired t-test. Switching to LLM-defined difficulty leads to higher accuracy gains both in models and datasets (Figure \ref{fig:best_accuracy_improvements}). Model-wise, TinyLlama-1.1B gave the highest accuracy gains in models (1.40\% and 1.44\%) in both data labelling scenarios. 
% Among the three data labelling scenarios, all learning strategies achieved a positive accuracy gain over Random Shuffle.
% Following the definition of maximum accuracy gain in Section~\ref{sec:evaluation}, the average maximum accuracy gain was 0.94\% across models and 1.02\% across datasets, both achieved with LLM-defined difficulty  (Table \ref{tab:best_accuracy_improvements}). 

\paragraph{Generalisation of human-inspired learning strategies} 
\label{sec:results_2}
% RESULTS PARAGRAPH: short and to the point, no discussion
% Table \ref{tab:all_combined_accuracy} shows that the best strategy is quite varied across models and datasets.
Table \ref{tab:all_combined_accuracy} presents the accuracy scores for all learning strategies, averaged across either datasets or models, with a total of 14 models and datasets shown in the table columns. Overall, Curriculum Learning was the top strategy in 5 out of 14 cases, followed by Interleaved Learning (4 cases) and Interleaved Curriculum (3 cases). Notably, Interleaved Learning consistently outperformed Random Shuffle across all models and datasets in both labelling scenarios (Table \ref{tab:accuracy_human_human} and \ref{tab:accuracy_llm_human}). However, when switching to LLM-defined difficulty, the best strategy shifts from Interleaved Learning to Interleaved Curriculum (Table \ref{tab:all_combined_accuracy}). 
% with the highest mean accuracy gains 

Moreover, there is no single strategy performing the best across all models in either data labelling scenario. For example, with human-defined difficulty (Table \ref{tab:accuracy_human_human}),  while Interleaved Learning consistently outperformed Random Shuffle, it wasn’t always the best for specific models. Curriculum Learning was most effective for Llama 2 7B and Llama 2 13B (up to +1.11\%) but under-performed compared to Random Shuffle on the other two models (up to -0.66\%). This suggests that strategies effective for one model may not generalise well to others, a trend also seen across datasets.

We also experimented with automating question category labelling using text clustering to group semantically similar questions. Switching to clustering-based question categories yielded accuracy gains at a similar scale (best accuracy gain of 1.77\% across models and 1.15\% across datasets), with a similar pattern observed, where no single strategy consistently outperformed the others across models or datasets (Appendix \ref{appendix:clustering}).
% % \paragraph{Impact of automated labelling on question categories}

% highlighting the need for caution to generalise the effects of strategies across different contexts based on limited results.

% Among the four models, three different best learning strategies were identified, each achieving maximum accuracy gains for one or two models. 

\paragraph{Performance of LLM-defined difficulty} 
\label{sec:cl_gains_results}
% RESULTS PARAGRAPH: short and to the point, no discussion
% With human-defined difficulty and pre-existing categories, only Interleaved Learning showed a noticeable accuracy improvement (+0.66) over Random Shuffle (Figure \ref{fig:three_results_comparison}a). 
% : Curriculum Learning (+0.05 to +0.42), Blocked Curriculum (+0.04 to +0.37) and Interleaved Curriculum (+0.07 to +0.84) 
% (+0.18 to +0.92)
% +0.16 to +1.81
% -1.05 to -0.14
Switching from human-defined to LLM-defined difficulty increased accuracy scores across all three curriculum-based strategies (Figure \ref{fig:llm_diff_improvements}), with all gains significant at the 10\% level in a paired t-test. The largest improvements were seen in MedMCQA for Blocked Curriculum (+0.74\%) and Interleaved Curriculum (+1.65\%), and in MedQA for Curriculum Learning (+0.91\%). Additionally, fine-tuning Mistral 7B, our best-performing model, on the MedQA training set (11.4k data) using LLM-defined difficulty showed that curriculum learning outperformed all other strategies across all datasets (Appendix \ref{appendix:medqa_ft_results}). These findings suggest that using LLM-defined difficulty  can improve the performance of curriculum-based learning strategies over human labels.

% DISCUSSION PARAGRAPHS (2-3): highlight different take home messages + discuss the literature in the context of this particular result
% \paragraph{Improvements of LLM-defined difficulty for curriculum design} 

\section{Discussion}
%%%%%%%%%%%%%%%%%%%%%%%%%%%%%%%%%%%%%%%
% \section{Related Work}
% \paragraph{Data-efficient fine-tuning of LLMs}
% Data selection and data ordering are the two key approaches in data-efficient fine-tuning of LLMs. \citet{llm_data_efficient_pretrain} investigated fine-tuning techniques that use zero-shot reasoning to select high-quality training examples, ensuring diverse samples represent the data distribution. \citet{deft} employed unsupervised core-set selection to minimise data requirements while maintaining fine-tuning accuracy. \citet{skillit} proposed an ordered data sampling algorithm for efficient data ordering for training advanced language processing skills. Our study differs from these studies by focusing on human learning inspired data ordering methods to improve fine-tuning efficiency. 
% While these approaches focus on selecting high-quality data subsets, 

\paragraph{Level of improvements} The highest accuracy gains in our study (1.81\% and 1.44\%) align with previous findings on curriculum learning. \citet{curri_nlu} reported a 1.3\% improvement in natural language understanding tasks by using multiple teacher models to define question difficulty for BERT fine-tuning, while \citet{curri_reasoning} showed a 2\% gain in common sense reasoning with RoBERTa fine-tuned on fixed and adaptive curricula. Our results are lower than \citet{llm_curriculum_learning}, who reported gains of 3.28\% and 1.73\% on World Knowledge and Commonsense Reasoning benchmarks for Llama-13B using Interleaved Curriculum. 

Two key factors may explain the difference. First, \citet{llm_curriculum_learning} used a broader curriculum, covering subjects from secondary to graduate level, while we focused on a specialised graduate-level medical curriculum, with a narrower scope of content. Second, their approach used Bloom’s taxonomy to classify questions by distinct difficulty levels, whereas our dataset, with more semantically similar medical questions, exhibits subtler difficulty variations. These differences in curriculum scope and difficulty categorisation may account for the variation in performance gains. 

% , making the classification of questions into rigid difficulty categories less applicable

% The differences may stem from their broader curriculum, covering subjects from secondary to graduate school levels, and clearer difficulty distinctions based on Bloom’s taxonomy, whereas our focus on medical exams offers a narrower scope and less distinct difficulty categories.

% domain difference (general education vs medical)
% the larger number of data points they used to fine-tune

% \citet{curri_bad} investigated linguistically motivated curricula for BERT pretraining but found no significant improvements over unordered data. In contrast, \citet{curri_nlu} showed a 1.3\% performance boost in natural language understanding tasks by using multiple teacher models to define question difficulty during BERT fine-tuning. \citet{curri_reasoning} reported a 2\% improvement in common sense reasoning tasks when RoBERTa was fine-tuned using fixed and adaptive curricula from a teacher model. Similarly, \citet{llm_curriculum_learning} found that interleaving subjects in the curriculum with Llama 2-13B improved MMLU benchmark scores by 3\%. 

% DISCUSSION PARAGRAPHS (2-3): highlight take home messages + discuss the literature in the context of this particular result
% \paragraph{Modest improvement of human-inspired learning strategies over Random Shuffle}
\paragraph{Limited generalisation}
Most previous studies demonstrated the effectiveness of curriculum learning using a single model, showing consistent improvements across several benchmarks \cite{curri_nlu, curri_reasoning, llm_curriculum_learning}. However, these results are insufficient to extrapolate that curriculum learning generalises well to other models, even though it is often assumed in these studies. Our findings suggest that a strategy effective for one model does not generalise to others,  emphasising the need for caution when generalising results and drawing conclusions from limited model-specific evidence. This limited generalisation across models and datasets may be due to variations in how different LLMs perceive question difficulty and inconsistencies in question categories across medical datasets.

% This contrasts with the results of Lee et al. \cite{llm_curriculum_learning}, where they found Interleaved Curriculum was consistently the best-performing strategy across multiple datasets. This discrepancy may be due to differences in the experimental design, as discussed in Section \ref{sec:results_1}. Although our results show that on average Interleaved Curriculum achieved the highest accuracy gain (+0.66) compared to other strategies (Figure \ref{fig:three_results_comparison}a), the margin of improvement compared to Lee et al. \cite{llm_curriculum_learning} was slightly smaller, who observed an average improvement of +1.85 across datasets.

\paragraph{Benefits of LLM-defined labels} Our results show that using LLM-generated difficulty modestly improves the performance of curriculum-based learning strategies. This aligns with previous studies, which found that language-model-ranked difficulty led to consistent accuracy gains in curriculum learning across benchmarks \cite{curri_reasoning, curri_nlu}. Unlike those studies, which first fine-tuned models with randomly shuffled data before ranking difficulty, we measured the difficulty directly, leveraging the pre-training knowledge of LLMs to define the labels. This demonstrates the potential of LLM-defined difficulty labels as a cost-effective alternative to human labels for optimising fine-tuning order in curriculum-based learning.

\section{Limitations}
% First, the five-time repetition we ran for each model-data combination gave dependent samples, which may make it less precise to establish reliable confidence intervals and statistical testing; aggregating these dependent samples into independent observations would also lead to sample sizes too small for accurate statistical analysis. 

% \paragraph{Results depend on model choices} Our evaluation results rely on the choices of LLMs and their pre-training medical knowledge. Future experiments could explore larger models (e.g., 70B parameters) and specialized medical LLMs to assess how model size and pre-training affect the impact of learning strategies.
%  Similarly, the results for clustered categories heavily depend on the choice of clustering algorithm and selected hyperparameters.

\paragraph{Fine-tuning data size may affect performance} The relatively small size of the LEK dataset, due to the limited human labels on medical questions, may restrict the visibility of certain effects of learning strategies, especially those that only emerge with larger data. For example, the benefits of Interleaved Learning might become only apparent over longer revision intervals, which our dataset might not fully capture. Similarly, the relative narrow range of question difficulties in the LEK dataset may limit the effectiveness of curriculum-based learning. 
% With LLM-defined difficulty labels, one could explore a medical curriculum that includes a broader spectrum of questions, from fundamental medical concepts to advanced-level knowledge, to test the effects of learning strategies more comprehensively. 

% Although our preliminary results with fine-tuning on the MedQA training set (11.4k observations) showed similar scale accuracy improvements (up to 2\%) on medical datasets using both Llama 7B and Mistral 7B models. 

% Absolute reproducibility hard to guarantee - the fine-tuned results may vary by different GPU models. Even with deterministic settings, the order of floating-point operations might differ slightly leading to small numerical differences.

\paragraph{Alternative notions for defining LLM-based difficulty} The LLM-defined question difficulty was labelled based on the average accuracy on questions from ensemble LLMs, which may not fully capture all aspects of difficulty. Alternative measures, such as topic-specific perplexity to assess topic familiarity or the average language modeling loss on sequences as an indicator of LLMs' pre-training medical knowledge, could be explored further.

\paragraph{Limited exploration of fine-tuning methods} Our study focused exclusively on supervised fine-tuning using the QLoRA approach. Future work could explore the effects of other fine-tuning methods, such as domain-adaptive pretraining (DAPT) \cite{dapt}, continual learning \cite{continual_learning}, or adapter-based fine-tuning \cite{adapter_fine_tuning}.

%\paragraph{Evaluate on more medical questions spanning a bigger curriculum}
% Future research could explore a medical curriculum that encompasses a broader spectrum of questions, spanning from fundamental medical concepts to advanced-level knowledge, to test the effects of curriculum-based learning more comprehensively. The range of categories and question proportions in each category could be controlled to reveal how much these factors affect the performance of category-based learning strategies.  

%\paragraph{Alternative clustering methods}
% Designing a careful data sampling strategy with clustering algorithms could lead to improved LLM performance. For example, Shao et al. \cite{clustering_lm_training} proposed ClusterClip Sampling, which balances common and rare samples during language model training based on clustered data distribution, resulting in an improvement 1\%-2\% over randomly sampled data. Therefore, future experiments could explore alternative clustering algorithms for category labelling with balanced data sampling to improve model performance. 

% \paragraph{Temporal dynamics of human-inspired fine-tuning}
% Lastly, investigating the fine-grained temporal dynamics of fine-tuning—such as whether easy questions are answered correctly first and how the distribution of correct answers evolves throughout the process—could provide valuable new insights into the mechanisms of human-inspired fine-tuning.

\section{Conclusions}
% summary
We evaluated human-inspired learning strategies for fine-tuning LLMs in medical question answering, with the evaluation conducted across four dimensions: learning strategies, models, datasets, and data labelling scenarios. Our findings show moderate improvements (up to 1.81\%) in fine-tuning with those strategies, suggesting some transferability of human learning behaviours to LLMs for more data-efficient tuning. However, the best learning strategy varied significantly across model-dataset combinations, with no single strategy consistently outperforming others, indicating limited generalisation of their effects across contexts. Additionally, we find that using LLM-defined difficulty measures provide moderate accuracy improvements over human-defined difficulty, highlighting their potential as a cost-effective alternative to human labels for optimising fine-tuning.

% suggesting caution when generalising their effectiveness based on limited model-specific evidence.

\newpage
% \printbibliography
% \bibliographystyle{unsrtnat}
\bibliographystyle{plainnat}
\bibliography{references.bib}

\newpage
\appendix

\section{Mean accuracy gains across data labelling scenarios}
\label{appendix:ft_results_per_dataset}
\begin{figure*}[h!]
\centering
\includegraphics[width=0.9\textwidth]
{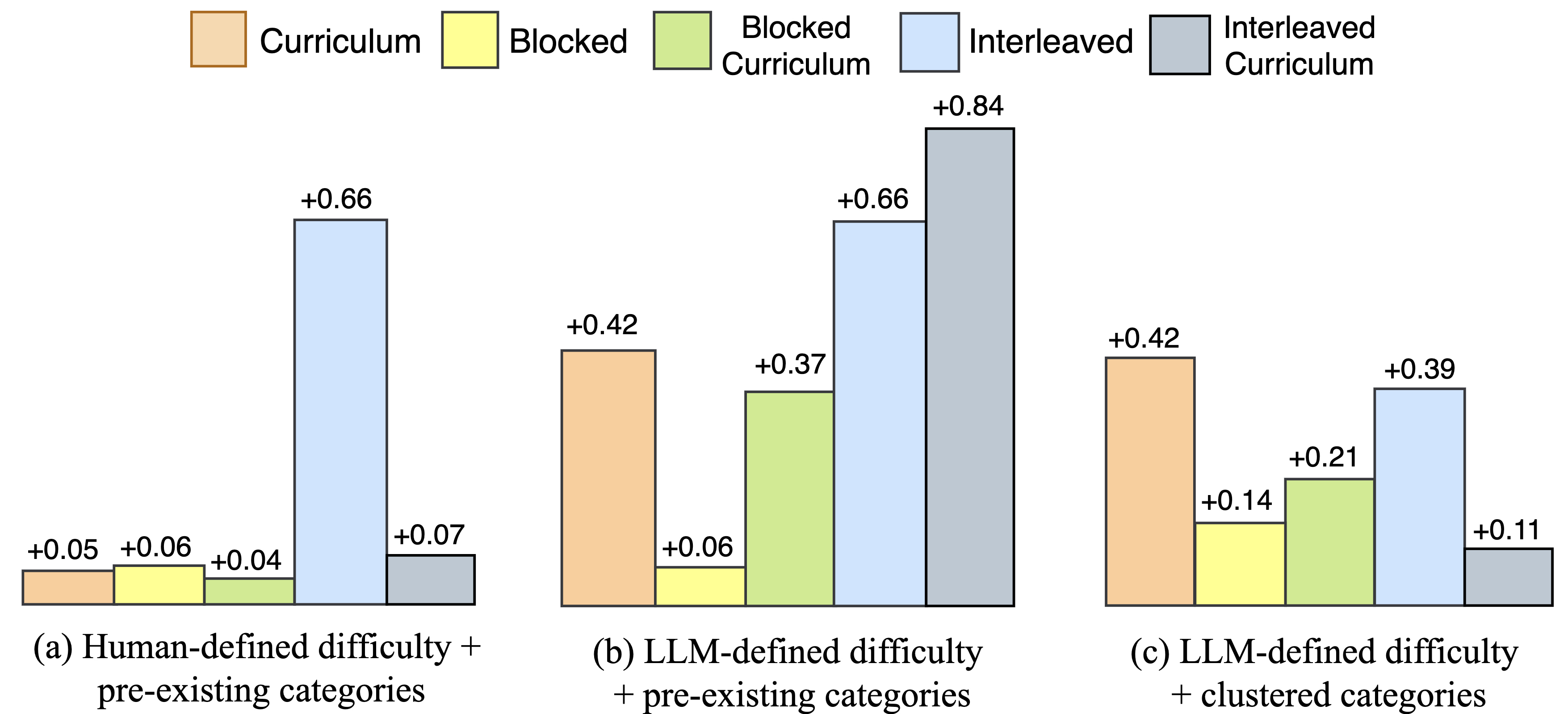}
% {Figures/results_fg_three_graphs.png}
\caption{\textbf{Mean accuracy gains for the learning strategies.} Each bar plot shows the mean accuracy gains (in \%) of the learning strategies when averaged across model-dataset combinations. Figures (a)-(c) represent three data labelling scenarios.}
\label{fig:three_results_comparison_datasets_old}
\end{figure*}

In this section, we present detailed results on the mean accuracy gains (in \%) of learning strategies across three data labelling scenarios: (a) Human-defined difficulty with human-defined categories, (b) LLM-defined difficulty with human-defined categories, and (c) LLM-defined difficulty with clustered categories, as shown in Figure \ref{fig:three_results_comparison_datasets_old}.

\section{Hyperparameters for fine-tuning}
\label{appendix:hyperparameters}
In this section, we present the hyperparameters for fine-tuning LLMs. The fixed hyperparameters for QLoRA were set as follows: 
\(r=16\), \(\alpha=64\), with a dropout of 0.1. AdamW was used as the optimizer, and the learning rate decay followed a linear schedule with no warmup steps. The maximum sequence length was set to 512. Table \ref{tab:model_hyperparameters} details the model-specific hyperparameters selected via grid search for each model on the Random Shuffle baseline. The grid search ranges were: learning rate in [5e-4, 1e-4, 5e-5, 1e-5, 5e-6, 1e-6, 5e-7, 1e-7], batch size in [4, 8, 16], and gradient accumulation steps in [1, 2, 4]. 
% For fine-tuning Mistral 7B on the MedQA training set (Appendix \ref{appendix:medqa_ft_results}), we changed the learning rate to 1e-7 and kept the same batch size and gradient accumulation step as shown in Table \ref{tab:model_hyperparameters}. 

% \vspace{0.1cm} % Adds space before the table

% model-varying fine-tuning hyperparameters
\begin{table}[h!]
\centering
\caption{\textbf{Model-specific hyperparameters for fine-tuning}. For each model, the hyperparameters were selected by grid search on the Random Shuffle baseline. Abbreviations: \textit{Grad accum.} = gradient accumulation steps. \textit{Tiny} = TinyLlama.}
\resizebox{0.6\textwidth}{!}{ % adjust table horizontal length
\begin{tabular}{|l|l|l|l|l|}
\hline
\textbf{}              & \textbf{Tiny} & \textbf{Llama 2} & \textbf{Llama 2} & \textbf{Mistral} \\ 
                       & \textbf{1.1B}      & \textbf{7B}     & \textbf{13B}    & \textbf{7B}      \\ \hline
\textbf{Learning rate} & 5e-4               & 5e-5            & 1e-4            & 1e-4             \\ \hline
\textbf{Batch size}    & 16                 & 4               & 4               & 4                \\ \hline
\textbf{Grad accum.}    & 1                  & 2               & 2               & 2                \\ \hline
\end{tabular}
}
\label{tab:model_hyperparameters}
\end{table}

\section{Automatic labelling on question categories}
\label{appendix:clustering}
% \paragraph{Clustering-based question categories}
In this section, we detail the process of automating question category assignment using text clustering to group semantically similar questions. First, we applied BioMedBERT sentence embeddings \cite{pubmedbert} to the question context and answers. We then used Uniform Manifold Approximation and Projection (UMAP) \cite{umap} for dimensionality reduction, followed by Hierarchical Density-Based Spatial Clustering of Applications with Noise (HDBSCAN) \cite{hdbscan} for clustering. HDBSCAN was chosen for its ability to handle noisy data and form clusters with variable densities without requiring a pre-defined number of clusters \cite{density_based_clustering}. Table \ref{tab:accuracy_llm_cluster} shows the accuracy scores of learning strategies with LLM-defined difficulty and clustered categories. Table \ref{tab:clustering_hp} shows the hyperparameters for clustering with UMAP and HDBSCAN for automated category labelling. The hyperparameters of UMAP and HDBSCAN were optimised using Bayesian optimisation to minimise the number of data points with low cluster probability (below 5\%). 
% It also specifies the hyperparameter ranges for Bayesian Optimisation. 
% The noise points identified by HDBSCAN were treated separately as an additional block in Blocked Learning. 

\begin{table}[ht]
    \centering
    \caption{\textbf{The accuracy scores (in \%) for learning strategies averaged across datasets or models, using LLM-defined difficulty and clustered categories as labels.} The best strategy still varies across models and datasets. In the \textit{Models} columns, scores are averaged over three datasets; in the \textit{Datasets} columns, scores are averaged over four models. \textit{Tiny} = TinyLlama.}
    \label{tab:accuracy_llm_cluster}
    \begin{tabularx}{0.9\textwidth}{|l|p{1.15cm}p{1.15cm}p{1.15cm}p{1.1cm}|p{1.15cm}p{1.15cm}p{1.0cm}|}
    \hline
    \multicolumn{1}{|c|}{\textbf{Strategy}} & \multicolumn{4}{c|}{\textbf{Models}} & \multicolumn{3}{c|}{\textbf{Datasets}} \\
    & \textbf{Tiny 1.1B} & \textbf{Llama2 7B} & \textbf{Llama2 13B} & \textbf{Mistral 7B} & \textbf{LEK} & \textbf{Med MCQA} & \textbf{Med QA} \\ \hline
    Random Shuffle & 20.40 & 38.71 & 42.57 & 47.97 & 43.55 & 36.28 & 32.40 \\
    Curriculum        & 20.88 & \textbf{39.21} & 42.82 & \textbf{48.39} & \textbf{44.36} & 36.86 & 32.26 \\
    Blocked               & 20.95 & 38.23 & \textbf{43.09} & 47.94 & 43.22 & 36.77 & \textbf{32.67} \\
    Blocked Curri.     & 21.50 & 38.39 & 43.00 & 47.62 & 43.12 & \textbf{37.43} & 32.32 \\
    Interleaved            & \textbf{22.17} & 38.23 & 43.03 & 47.77 & 43.61 & 37.39 & 32.41 \\
    Interleaved Curri. & 20.74 & 38.45 & 43.01 & 47.87 & 43.33 & 37.34 & 31.88 \\ \hline
    \end{tabularx}
\end{table}

% clustering hyperparameters
\begin{table}[h!]
\centering
\caption{\textbf{Hyperparameters for clustering with UMAP and HDBSCAN.} The \textit{Range} row specifies the range for hyperparameter search in Bayesian Optimisation, the \textit{Set} row specifies the hyperparameter value selected.}
\resizebox{0.62\textwidth}{!}{
\begin{tabular}{|ll|ll|ll|}
\hline
\multicolumn{2}{|l|}{}                               & \multicolumn{2}{c|}{\textbf{LEK}}                                & \multicolumn{2}{c|}{\textbf{MedQA}}           \\
\multicolumn{2}{|l|}{}                               & \multicolumn{1}{l}{Range} & Set                                 & \multicolumn{1}{l}{Range}            & Set         \\ \hline
\multicolumn{1}{|l|}{\multirow{2}{*}{\textbf{UMAP}}} & \text{Number of} & \multicolumn{1}{l}{{[}8, 20{]}}    & 15    & \multicolumn{1}{l}{{[}5, 30{]}}      & 5    \\ 
\multicolumn{1}{|l|}{}                               & \text{Neighbours}  & \multicolumn{1}{l}{}             &       & \multicolumn{1}{l}{}                 &    \\ \cline{2-6} 
\multicolumn{1}{|l|}{}                               & \text{Number of} & \multicolumn{1}{l}{{[}3, 15{]}}    & 5     & \multicolumn{1}{l}{{[}3, 20{]}}      & 17   \\ 
\multicolumn{1}{|l|}{}                               & \text{Components}  & \multicolumn{1}{l}{}             &       & \multicolumn{1}{l}{}                 &    \\ \hline
\multicolumn{1}{|l|}{\textbf{HDBSCAN}}               & \text{Minimum} & \multicolumn{1}{l}{{[}25, 35{]}}     & 25    & \multicolumn{1}{l}{{[}200, 250{]}}   & 202  \\ 
\multicolumn{1}{|l|}{}                               & \text{Cluster Size}  & \multicolumn{1}{l}{}           &       & \multicolumn{1}{l}{}                 &    \\ \hline
\end{tabular}
}
\label{tab:clustering_hp}
\end{table}

\section{Fine-tuning on MedQA training set}
\label{appendix:medqa_ft_results}
In this section, we present the results of fine-tuning the MedQA training set (11.4k data points) using the Mistral 7B model, our best-performing model. We employed LLM-defined difficulty and clustered categories as data labels, as the MedQA dataset lacks pre-existing human-defined medical categories \cite{medqa}. The LLMs used to compute difficulty measures were Mixtral 8x7B \cite{mixtral}, Meditron 70B \cite{meditron}, Llama 2 70B \cite{llama2}, and Jamba \cite{jamba}. Table \ref{tab:medqa_all_scores} shows that Curriculum Learning consistently outperformed other strategies across all three datasets, with the highest average accuracy gain over Random Shuffle (+0.70\%).

% \vspace{0.1cm} % Adds space before the table

% ------------------Mistral-7B--------------------
\begin{table}[ht]
\centering
\caption{\textbf{Accuracy scores of Mistral 7B fine-tuned on MedQA training set}. The accuracy scores (in \%) were computed using LLM-defined difficulty and clustered categories as data labels.}
% Abbreviations: \textit{Blocked Curri.} = Blocked Curriculum, \textit{Interleaved Curri.} = Interleaved Curriculum.
\resizebox{0.57\textwidth}{!}{ 
\begin{tabular}{|l|c|c|c|c|}
\hline
\textbf{Strategy} & \textbf{LEK} & \shortstack{\textbf{Med} \\ \textbf{MCQA}} & \textbf{MedQA} & \textbf{AVG} \\ \hline
% No Fine-tuning & 38.95 & 38.75 & 50.46 & 42.72 \\ \hline
Random Shuffle & 44.38 & 41.67 & 50.57 & 45.54 \\
Curriculum & \textbf{45.40} & \textbf{42.19} & \textbf{51.14} & \textbf{46.24} \\
Blocked & 44.76 & 41.70 & 50.71 & 45.72 \\
Blocked Curri. & 44.64 & 41.89 & 50.64 & 45.72 \\
Interleaved & 44.65 & 41.75 & 50.87 & 45.76 \\
Interleaved Curri. & 44.92 & 42.06 & 50.73 & 45.90 \\ \hline
\end{tabular}
}
\label{tab:medqa_all_scores}
\end{table}

\end{document}